\documentclass[letterpaper, 10 pt, conference]{ieeeconf}  % Comment this line out if you need a4paper
\IEEEoverridecommandlockouts                              % This command is only needed if 
\title{\LARGE \bf
Exploiting Deep Semantics and Compositionality of Natural Language for Human-Robot-Interaction}
\author{Manfred Eppe$^{1}$, Sean Trott$^{1}$, Jerome Feldman$^{1}$% <-this % stops a space
\thanks{$^{1}$International Computer Science Institute, Berkeley, USA
        {\tt\small \{eppe, seantrott, feldman\} @icsi.berkeley.edu}}%
}
\usepackage{etex}
\usepackage{times}
\usepackage{booktabs}
\usepackage{amsmath}
\usepackage{amssymb}
\usepackage{mathtools}
\usepackage{wrapfig}

\usepackage{tikz}
\usepackage{tikz-qtree}
\usetikzlibrary{arrows,shapes,positioning,calc,intersections}

\usepackage{url}

\usepackage{times}

\usepackage{colortbl}

\usepackage{verbatim}

\usepackage{caption}
\usepackage{graphicx}

\usepackage{framed}

\usepackage{proof}

\usepackage{mathrsfs}

\usepackage{paralist}

\usepackage{thmtools}

\usepackage{longtable}

\colorlet{darkblue}{blue!50!black}
\colorlet{hlinkcolor}{darkblue}
\usepackage[colorlinks=true,linkcolor=hlinkcolor,urlcolor=hlinkcolor,citecolor=hlinkcolor]{hyperref}

\newcommand{\co}[1]{\ensuremath{\text{\texttt{#1}}}}

\newcommand{\citet}[1]{\cite{#1}}

\colorlet{lGray}{gray!50!white}
\colorlet{lBlue}{blue!20!white}
\colorlet{yGreen}{white!50!black!20!yellow}

\newcommand{\ra}[1]{\renewcommand{\arraystretch}{#1}}

\begin{document}
\maketitle
\thispagestyle{empty}
\pagestyle{empty}
\begin{abstract}
We develop a natural language interface for human robot interaction that implements reasoning about deep semantics in natural language. To realize the required deep analysis, we employ methods from cognitive linguistics, namely the modular and compositional framework of Embodied Construction Grammar (ECG) \cite{Feldman2009}.
Using ECG, robots are able to solve fine-grained reference resolution problems and other issues related to deep semantics and compositionality of natural language. 
This also includes verbal interaction with humans to clarify commands and queries that are too ambiguous to be executed safely. 
We implement our NLU framework as a ROS package and present proof-of-concept scenarios with different robots, as well as a survey on the state of the art. 
\end{abstract}
%
%!TEX root = NLU-Robots.tex
\section{INTRODUCTION}
Robots are becoming more and more our companions and co-workers. 
Hence, a lot of effort has been made to develop methodologies to make the interaction between humans and robots seamless, intuitive and uncomplicated. 
This includes interface devices that rely on pointing and other gestures, as well as algorithms that predict human behavior to produce appropriate non-interfering robot behavior. 
Surprisingly however, only relatively shallow work has been conducted with one of the most natural of all interaction methods with robots, namely Natural Language. 

Of course, there exist several robots and other virtual agents that are equipped with some kind of natural language interface. 
Examples of current commercial products are Apple's Siri, Microsoft's Corona, Google Now, and the social multimedia robot \emph{Jibo} \cite{jibo2015}. 
A problem with these products is that they rely mostly on learned or predefined keyword-based input templates that trigger certain actions. 
This limits the subset of language that these products can interpret.
In the literature, such shallow approaches are usually referred to as \emph{Natural Language Processing} (NLP). 
NLP systems are appropriate when the results are for human consumption rather than direct action. For robots, automatic cars, and other autonomous physical systems, however, this is usually not sufficiently reliable, and their traditional safety layers can not cover all possible consequences of misunderstanding. 

The underlying issue with NLP approaches is that they can not analyze sentences semantically. 
This is very important to understand metaphors, conditional statements, and to reliably clarify ambiguous references. 
As an example, consider a laboratory robot assistance situation. 
Neither Siri, Corona, nor Google Now are able to understand sentences like 
``If there is an empty test tube to the left of the bottle with sulfuric acid, please pour 10 ml ammonia in it.''
Cognitive linguistics clearly tells us that the ``it'' refers to the ``empty test tube'', and not the ``sulfuric acid'', but it is not trivial to infer this computationally.
If the robot accidentally misinterprets this sentence, and pours the ammonia into the acid, this will cause an undesired strong and dangerous reaction.%
\footnote{In fact, there are two ways to resolve references, which are usually applied in combination. One is the interpretation of sentence structure, and the other is applying commonsense background knowledge. In this paper we focus on the former, but note that using semantic knowledge for disambiguation has been demonstrated for a limited robotics domain \cite{Trott2015}.}
It is well known that approaches that consider deep language semantics in form of abstract grammatical constructions and concepts are capable of analyzing statements like the one above reliably (see e.g. \cite{Goldberg1995}). 
We talk about \emph{Natural Language Understanding} (NLU) when we refer to such systems.

Fortunately, robotics offers many properties that are a good fit for NLU systems. Most importantly, one usually designs a language interface for a robot that is supposed to work in a closed domain, such as assisted living, household, or disaster response. 
This can be exploited in that learning is not required, and one can focus on reliable grammar based approaches.
In this work, we present such a grammar based approach for NLU in HRI, and formulate our straight-forward hypothesis as follows:

\emph{Incorporating cognitive theories and linguistic expert knowledge leads to significant improvements in language understanding for Human-Robot-Interaction.}

Surprisingly, this simple hypothesis has not yet been deeply addressed in robotics (see Sections \ref{sec:prelim:survey} and \ref{sec:eval:related_work} on related work and evaluation). We believe that the lack of mature computational systems for deep NLU is a major reason for this shortcoming. 
In our approach, we use an \emph{Embodied Construction Grammar} (ECG) \cite{Feldman2009} analyzer to capture deep semantics in natural language by pairing syntax with semantics and universal embodied concepts. 
An advantage of ECG vs. other cognitive linguistic frameworks is its modularity. 
There is a universal core set of constructions and conceptual schemas, developed by experts from cognitive linguistics. 
These capture modalities, spatial relations, temporal expressions, actions, etc. 
The core set is common across domain-specific grammars, while the domain specific grammars themselves are usually relatively small, requiring only little knowledge about linguistics. 
The conceptual schemas in this core set are neurologically and cognitively motivated, and therefore largely language independent \cite{Feldman2009,Lakoff1999}.  

In this paper, we describe how we connect the ECG language analyzer to the Robot Operating System (ROS) \cite{ros2015} in an effort to make it accessible to a wide range of robotic platforms.
We also review the core grammar set to fine-tune it for robotic applications. We do not claim to solve all linguistic problems that exist, but we do claim that our modular language understanding framework does better than the state of the art for an important subset of linguistics problems, which we illustrate in Table \ref{tab:comparisonOfFeatures}. 
We also do not claim to have solved the full pipeline of the IROS 2016 theme, ``Road  to  companionship  with  intelligent  robots  in  everyday  life  and workspaces'', which also includes speech recognition, vision, action planning and other problem solving. 
Instead, we leave these problems to the respective experts, and assume that they are sufficiently solved for our purposes.
However, we do claim that our approach is an important part of that pipeline, and that it contributes to an key aspect of robot companionship, namely exploiting deep semantics in natural language to facilitate understanding.

%
%!TEX root = NLU-Robots.tex
\section{PRELIMINARIES AND RELATED WORK}
\label{sec:prelim}

\subsection{What Makes Natural Language Understanding Hard}
\label{sec:prelim:nat-lang-probs}
Natural Language Understanding is generally still unsolved. However, we can exploit the fact that robotics usually requires only a finite, albeit large, domain description, so that grammar-based approaches with a controlled language subset are applicable. Under these circumstances, one can constrain the focus on the following specific solvable issues:

\paragraph{Compositionality and the combinatorial explosion of form and meaning}
\label{sec:prelim:nat-lang-probs:compositionality}
Word meaning is usually heavily overloaded. As an example, consider motion words like ``move'' which has a very different semantics in, e.g., the transitive expression ``move the table'' vs. intransitive ``move to the table''.
Often, the appropriate meaning is determined by the construction, in this case transitive vs. intransitive, in which it appears \cite{Goldberg1995}.
This does not only lead to a combinatorial explosion of form and meaning, but it also allows one to invent novel and uncommon word meanings spontaneously. For example, the expression ``to sneeze the napkin off the table'' imposes a movement semantics on the verb ``sneeze'', which is rare but absolutely legitimate.

Under the condition that enough data is available, learning based approaches are attractive to solve the combinatorial problem  \cite{Bastianelli2013}. 
However, they can not deal with the novel inventions depicted above. 
Also, as the language subset they are trained on grows larger, learning based language interpretation for secure robot operation becomes less precise and deterministic.
To understand a large language subset, including novel and previously unheard expressions, while also being robust to misinterpretation, one requires a more sophisticated and modular way to compose grammatical primitives. 
This is exactly what \emph{construction grammars} do \cite{Goldberg1995}, as we describe in Section \ref{sec:prelim:ecg}.

\paragraph{Reference resolution and ambiguity} Reference resolution means to identify an object in the real world that is referred to by a pronoun, as e.g. in the sentence ``If there is a cup on the dining table, please bring it to me''. 
Cognitive linguistics tells us that clearly the ``it'' refers to the cup and not to the table, because the ``cup'' is the head noun and the dining table is the subordinate description \cite{Goldberg1995}. 
In contrast, consider ``If there is a table under the cup, please bring it to me''. 
Here, the ``it'' clearly refers to the table. 
Note that in addition it carries the implicit information that the speaker expects the hearer to be capable of carrying the table. 
Such deep semantics can be exploited for HRI. 

In general, reference resolution consists of two steps. 
Firstly, \emph{anaphora resolution} identifies the noun to which a pronoun refers within a sentence. Secondly, \emph{grounding} identifies the object in the real world to which the noun in the sentence refers. 
% In the general case, ambiguities do not necessarily involve a pronoun. For example, consider the following simple conversation, where ambiguity is a result of ambiguous grounding possibilities: ``Please bring me the pen on the table!'' -- ``There is a red one and a blue one, which one?'' -- ``The blue one.'' 
In cases where grounding can not disambiguate the reference, clarification dialogs are required. 
As an example, consider the scenario in Sec. \ref{sec:eval:scenarios}, where the DARwin-OP needs to resolve an ambiguity to decide which object to pick up (in this case the blue marker and not the red one).

\paragraph{Conditionals} Conditionals are another crucial building block of verbal articulation, which usually follow an \emph{if-then-else} pattern, with the \emph{else} being optional. 
For illustration consider our scenario with the PR2 robot in Sec. \ref{sec:eval:scenarios}

\paragraph{Erroneous input and ungrammatical sloppy language} Input text is often erroneous in different ways. 
For example, when given as a transcript from speech recognition systems, as in ``Please bring me the grass'' instead of the correct ``Please bring me the glass.'' Furthermore, users tend to often use jargon and ungrammatical or sloppy language. 

\paragraph{Disfluency analysis and repair} Humans often abandon sentences and words midway during the speaking, switching to other conversation segments or correcting sentences on the fly. 
This makes language disfluent and requires repair. 
For example, ``There is a blue kit, erm, I mean a blue box, at the end of the table'' can be simplified to ``There is a blue box at the end of the table'' \cite{Cantrell2010}.

\paragraph{Indirect assertions through relative clauses and appositions} Relative clauses and appositions are often used for indirect assertions of world properties. 
As an example consider: ``The noodles, which are still in the pot on the stove, must be wasted by now; bring them to the trash!''
This sentence encodes the information about the location of the noodles in an embedded relative clause, which needs to be interpreted correctly before the command can be executed.

\paragraph{Modalities} Modalities are used to express ability, knowledge, temporality and other mental attitudes. For example, ``Are you able to order pizza?'' is not a question about the pizza but about the ability to order pizza (imagine an assistance robot that is connected to an online pizza service).

\paragraph{Indirect speech acts} These are often used for politeness and sociality. For example, instead of saying ``Open the window!'', one often uses the more polite ``Can you open the window?'' While this is literally a modality question about the hearer's abilities, a robot should instead treat it as a command. 
It is an issue to decide when a robot should interpret a modality literally or as indirect speech \cite{Williams2015}.

\paragraph{Interlocutor feedback} During a dialog, humans constantly give feedback to each other. For example, a hearer (H) often uses  ``[o]kay'' to notify the speaker (S) that he understood the last part of an instruction correctly. S: ``so turn right'' H:``kay'' S: ``and walk a little bit and turn right again''\cite{Cantrell2010}. Feedback can also be nonverbal (gaze, pointing), but that is out of our work's scope. 

\paragraph{Metaphor} Metaphor is much more than an instrument of poetry and arts. 
Humans frequently use metaphors in language, often even unconsciously, and there is significant evidence suggesting that metaphor plays a central role in abstract thought \cite{Lakoff1980}. 
Therefore, metaphor is also crucial to understand language, and there has been extensive work on the cognitive linguistics of metaphor \cite{Dodge2015}. 
However, this has not yet been applied to human robot interaction.

\paragraph{Integration of Robots with NLU systems}
Connecting a NLU system to a robotic platform is not trivial. Firstly, NLU is a problem which is completely independent from robotic problems like action and motion planning, so it should be independent from the robot that one wants to use. Therefore, one should integrate an NLU system into a robotic framework that is modular, such as ROS.
Secondly, it is problematic to interpret the semantic specification of the NLU analysis as robot commands. For example, after analyzing the sentence ``if there is tea on the table, please bring it to me'', one needs to find an appropriate representation that is suitable as input for a robot's problem solving mechanism. 
In the case of the above sentence, we have an epistemic planning problem with possibly incomplete world knowledge (see e.g. \cite{Eppe2013c}). 
The NLU system needs to communicate with the problem solver on an abstraction layer that is appropriate wrt. the underlying reasoning problems.

\subsection{Embodied Construction Grammar and Compositionality}
\label{sec:prelim:ecg}
\begin{figure}
\vspace{6pt}
\includegraphics[width=\columnwidth]{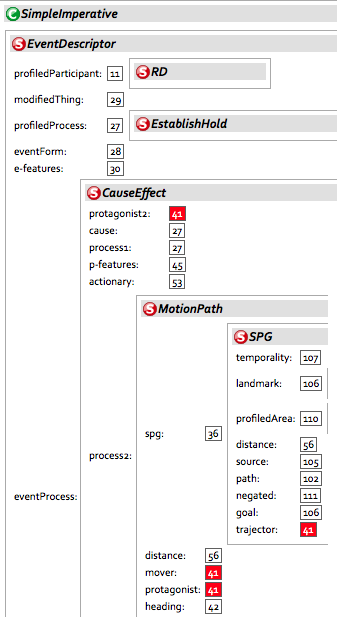}
\caption{Excerpt from SemSpec for utterance ``PR2, bring the soda can to the dining table!'' .}
\label{fig:semspec}
\vspace{-10pt}
\end{figure}
In linguistics, there are several kinds of compositionality. The most simple one is lexical compositionality, and it is the key feature of Chomsky's context-free grammars, which most existing NLU system for robotics use \cite{Pullum1982}. 
A problem with this approach is that it does not have adequate coverage, as described in Sec. \ref{sec:prelim:nat-lang-probs}, item a).
Construction grammar involves the pairing of form and meaning and allows one to model higher-level linguistic concepts, such as transitivity (``move the table'') and intransitivity (``move to the table''). This is done in an abstract way that is independent from the actual verb in the respective construction. 

Because such higher level constructions can be commonly used across all domains, the grammars remain compact, highly generalizable and productive. 
Hence, one can easily increase the size of the controlled language subset for the particular domain without much effort, while still having the advantage of being more reliable than systems based on learning.
Note also that higher-level linguistic concepts like transitivity are often similar or even identical for many languages, which makes it easier to develop a grammar for a new language by starting from another language. 

\emph{Embodied} Construction Grammar (ECG) \cite{Feldman2009} goes one step further and not only provides a way to compose constructions, but also concepts.
These concepts are neurologically and cognitively motivated in accordance with the established theory about \emph{image schemas} \cite{Lakoff1999}. 

As an example, consider the Semantic Specification (SemSpec) in Figure \ref{fig:semspec}, a reduced semantic analysis of the utterance ``PR2, bring the soda can to the dining table''. 
The motion verb ``bring'' triggers the \emph{EstablishHold} schema, which suggests that some \emph{protagonist} must grasp an \emph{actedUpon} object. 
The transitive construction (``bring the soda can'') followed by a prepositional phrase (``to the dining table'') triggers the \emph{CauseEffect} schema, in which the \emph{affectedProcess} is \emph{MotionPath}. This in turn evokes the \emph{Source Path Goal} (SPG) schema, which implies that there must exist some trajectory from a source location via a path to a goal location, and there must exist a trajector that follows this path. 
Conceptual compositionality is illustrated through co-indexing of roles both within and across schemas.
Note that the \emph{mover} role in the \emph{MotionPath} schema is co-indexed and identified with the id ``41'' in the \emph{trajector} role in the SPG schema, as well as the \emph{affectedEntity} role in the \emph{CauseEffect} schema. 
The SemSpec represents the conceptual network involved in understanding this utterance, and the relations between each component of this conceptual network.
A crucial feature of this approach is that schemas are language independent. 
For example, the concept of \emph{Source Path Goal} is universal in all languages and cognitively motivated \cite{Lakoff1999}.

% A technical description of the analysis system is provided in Section \ref{sec:sys_desc:ecg}. 
A computational framework that can analyze language according to the ECG theory has been developed over the last decade \cite{Bryant2008,Feldman2009a,Sinha2008,Khayrallah2015,Trott2015}.
The framework has been used in understanding and deep meaning of a wide range of complex constraints in multiple languages \cite{Khayrallah2015,Trott2015} beyond those described in this article.

\subsection{A Brief Survey on NLU for Robotics} 
\label{sec:prelim:survey}
In the following, we present a survey on related work on NLP and NLU for robotics. A summary of our findings can be found in the evaluation in Section \ref{sec:eval:related_work}, Table \ref{tab:comparisonOfFeatures}.

\smallskip
The work in \citet{Cantrell2010} is probably the most relevant for our approach. The authors describe the NLU module for the \emph{DIARC} HRI-System \cite{Scheutz2013}. They perform Wizard-of-Oz experiments \cite{Scheutz2008} to identify several issues that are crucial for \emph{natural} language understanding for HRI, especially when using speech input. 
	The authors build a system that can deal with a considerable part of the linguistic problems described in Sec. \ref{sec:prelim:nat-lang-probs}. 
	Specifically, their architecture is capable of speech recognition, semantic analysis, disfluency analysis and reference resolution.
	For the semantic analysis, the authors use Combinatory Categorical Grammar (CCG) and lambda conversions \cite{Steedman2000}. 
	A probabilistic extension, where the authors use a statistical Dempster-Shafer-theoretic approach to deal with natural language for robotics is presented in \cite{Williams2015}. This approach focuses on intention detection and indirect speech acts.
	It does not explicitly mention modalities, but modalities are obviously often used in indirect speech acts, e.g., in ``Do you know what time it is?''. 
	However, it remains unclear whether the authors' system can analyze modalities in the case where they are not used indirectly. Conditional statements are also not mentioned in the articles. 
	Grounding for reference resolution is possible, albeit it remains unclear how the anaphora resolution is performed.

\smallskip
The authors of \cite{Kruijff2007,Kruijff2010} focus on dialog. 
They represent the semantics of an utterance in a categorical modal-logical form, based on Combinatory Categorical Grammar (CCG) \cite{Steedman2000}.
	Their system can perform reference resolution and starts a clarification dialog if the reference is too ambiguous. For disfluency analysis, the authors use contextual knowledge to prime utterances. Verbal feedback with words like ``okay'' or ``fine'', can also be handled. 
	The authors present an implementation which they use to perform experiments with impressive results, but do not demonstrate how the system is connected to real or simulated robots in a modular manner. 

\smallskip
Another construction grammar which is also often used in Robotics is Fluid Construction Grammar (FCG) \cite{Steels2012,Steels2015,Spranger2015}. 
	For the most part, the goal of existing work around FCG in robotics is not Human-Robot-Interaction, but to provide a model for the evolution of language in robotic communities through so-called language games \cite{Steels1998a}. 
	However, the authors also present the \emph{Talking Heads Experiment}, which shows that the emergence of lexico-semantic models in robots can also be combined with human interaction, such that human language is acquired naturally by robots \cite{Steels2015}.
	The work on FCG has been extended in \cite{Beuls2012} to cope with erroneous input, disfluencies and repair. 

\smallskip
The approach of \cite{Bos2007} uses Discourse Representation Theory (DRT) \cite{Kamp2011} to capture semantics in natural language. An important aspect of DRT is that it can be compiled into first-order logical Discourse Representation Structure (DRS), which allows one to perform problem solving and higher level reasoning. 
	DRS is compositional, and it also allows for reference resolution and related problems that result from the deep semantics of natural language. 
	% First order logic can also naturally encode conditionals and provides a clean way to express statements and knowledge in a way that is accessible for a robotic agent. However, in the paper 
	The authors ``[...] used a generic domain-independent, but linguistically motivated grammar as a starting point'' \cite{Bos2007}, but it remains unclear how rich the grammar is and in how far it supports linguistic problems like conditionals, transitivity, and other advanced linguistic argument structure for reasoning. 

\smallskip
The information-theoretic probabilistic approach by \cite{Tellex2011,Tellex2012,Deits2012,Walter2013} seeks to minimize uncertainty of commands by asking questions that maximize the expected disambiguation. 
	Hence, the focus is on dialog and clarification. 
	The system is based on learning and so-called Spatial Description Clauses and Generalized Grounding Graphs \cite{Tellex2011}.
	% The work is extended in \cite{Walter2013} towards semantic learning of maps. 
	The constructional and conceptual compositionality aspect of language is not mentioned in the paper, and linguistic problems like conditionals, appositions, relative clauses, etc. are not specifically addressed. 

\smallskip
Recent work presented in \cite{Chai2014} focuses on the problem that humans and robots do not have a common perceptual ground, which is important for dialog. The problem is that the background knowledge and the object recognition capabilities of robots are far behind human level, so that grounding objects in the real world is difficult. 
% For example, a human might talk about a green cup, but the robot does not recognize the green cup. 
In this context, the authors investigate the additional effort that is required to establish a common perceptual ground between robot and human. The authors demonstrate their approach with an implementation and experiments on humanoid NAO robots. 

\smallskip
Impressive recent work by \citet{Barrett2015a} presents a supervised learning method to let a a robot learn word meaning for navigation purposes and spatial relations. 
	The authors emphasize their use of continuous domains, instead of using symbolic primitives like ``Drive to Location 1''. 
	% They use a corpus of 600 Examples that show videos of how a teleoperated robot is driving, and use supervised learning to process expressions like ``The robot went in front of the chair which is left of the table and behind the cone then toward the cone and right of the chair right of the cone.''. 
	% The language that the system can understand is limited to navigation, and it is unclear how scalable it performs for bigger domains. 
	The language model the authors use is based on a simple non-compositional grammar that can deal with relative sentences and some intermediate grammatical structure. 

\smallskip
\citet{Bastianelli2013} describe a method to interpret natural language using frame semantics \cite{Fillmore1985,Baker1998,Petruck1996}. 
Their framework allows them to correctly parse simple sentences like ``take the book on the table'' by mapping them to a form they call \emph{Abstract Meaning Representation} (AMR), which in turn can be mapped to robot commands. 
	The grammar they use is context-free.
	The authors elaborate their work in \cite{Bastianelli2014}, where they compare a grammar-based NLU system with a statistical learning one, and combine the two to obtain a hybrid system. 
	The authors do not demonstrate how they analyze conditionals, relative sentences, modalities and other rich linguistic structure. They handle disfluencies by adding wildcards for ``please'' or ``erm'' to their grammar. 
	A notable contribution is also the integration of different corpora, namely the RoboCup@Home \cite{Wisspeintner2009}, the Speaky4Robot \cite{Aiello2013} corpora, and another grammar generated corpus, which can be used for the evaluation of NLU systems \cite{Bastianelli2014a}. 
	However, a problem is that these corpora contain mostly very clean and simple sentences without conditionals, anaphora resolution or other problems that we describe in Section \ref{sec:prelim:nat-lang-probs}

\smallskip		
Another approach that focuses on navigation is presented in \citet{MacMahon2006}. 
	The authors use a corpus of natural language text to follow instructions, and show that their agent follows navigation commands nearly as precise as humans. However, the system is restricted to navigational language and does not use deeper semantics or compositional grammar. Accordingly, no deep semantic expressions like conditionals or metaphors in language is accounted for, and reference resolution is also not mentioned. 
	The authors perform experiments in a simulated environment. 

\smallskip	
The work presented in \citet{Matuszek2012} describe the learning of a parser to infer robot commands from natural language. 
	Towards this, the authors define a Robot Control Language (RCL) that is used to ground natural language in robot actions (specifically navigation actions). They learn a probabilistic version of Combinatory Categorial Grammar (CCG) and map it to RCL. 
	CCG features constructional compositionality, which would make it easy to also deal with conditionals or other rich structure in language, but we have not found the use of such structure in the paper.

%
%!TEX root = NLU-Robots.tex
\begin{figure}
\includegraphics[width=\columnwidth]{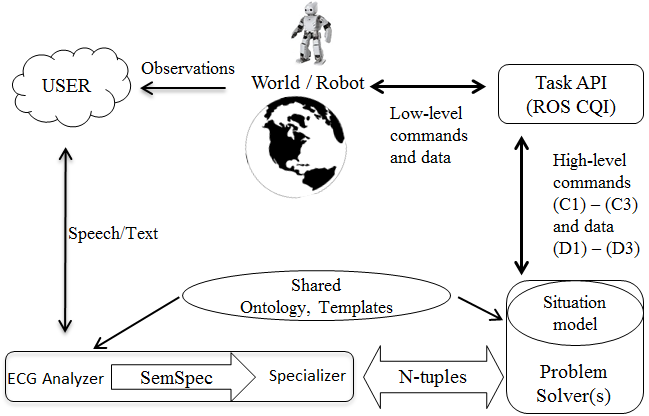}
\caption{ECG analysis architecture.}
\label{fig:diagram}
\vspace{-10pt}
\end{figure}

 \begin{figure}[!h]
 \vspace{6pt}
 \centering
 \includegraphics[width=.7\columnwidth]{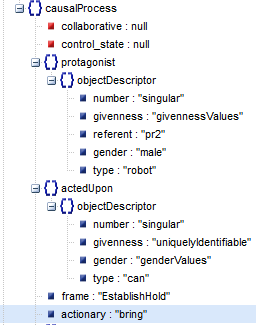}
 \caption{N-tuple excerpt for input utterance: ``PR2, bring the soda can to the dining table!'', Here, ``PR2'' is automatically  identified as protagonist in a causal process that executes a ``bring'' action upon an object of type ``can''.} 
 \label{fig:ntuple}
 \vspace{-10pt}
\end{figure}

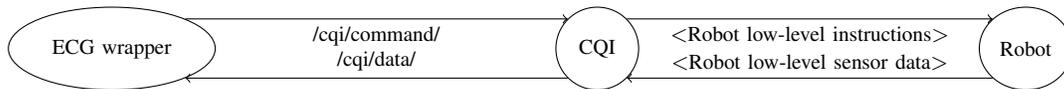
\begin{figure*}[ht!]
\vspace{6pt}
\centering
\begin{tikzpicture}
\footnotesize
\node[draw,ellipse, minimum size=1.1cm] (ECGW) {\begin{minipage}{50pt}\centering ECG wrapper \end{minipage}};
% \node[above = of ECGW,draw,xshift=100pt] (gazebo) {Gazebo};
\node[right = of ECGW,draw,ellipse, minimum size=1.1cm,xshift=100pt] (CQI) {CQI};
\node[right = of CQI,draw,ellipse, minimum size=1.1cm,xshift=100pt] (darwin) {Robot};
\draw[->] (ECGW.north east) -- (CQI.north west) node[midway,below]{/cqi/command/};
% \draw[->] (gazebo) -- (CQI) node[midway,right]{/gazebo/model\_states/};
\draw[->] (CQI.south west) -- (ECGW.south east) node[midway,above]{/cqi/data/};
\draw[->] (CQI.north east) -- (darwin.north west) node[midway,below]{$<$Robot low-level instructions$>$};
\draw[->] (darwin.south west) -- (CQI.south east) node[midway,above]{$<$Robot low-level sensor data$>$};
\normalsize
\end{tikzpicture}
\vspace{-5pt}
\caption{ECG-ROS communication flow}
\label{fig:ecg-ros-commstructure}
\vspace{-10pt}
\end{figure*}
\section{SYSTEM OVERVIEW}
This section describes our ECG analyzer and its integration with ROS. 
The overall system consists of two nested parts. 
Firstly, there is the ECG analyzer which takes natural language as input and gives high-level robot commands as output. 
Secondly, we describe how the ECG analyzer is nested and embedded within ROS.
We provide an additional illustration of the system in the supplementary video.

\subsection{The ECG semantic analysis framework}
\label{sec:sys_desc:ecg}
The semantic analysis is illustrated in Figure \ref{fig:diagram}.
Firstly, the ECG Analyzer \cite{Bryant2008} uses an ECG grammar to parses a sentence and perform a best-fit analysis which produces so-called \emph{Semantic Specifications}, or \emph{SemSpecs}. 
We depict a partial example of a SemSepc in Fig. \ref{fig:semspec}.
To transform the semantic specifications into actual robot commands, several intermediate steps are necessary. 
The second step is performed by the \emph{Specializer}, which crawls the SemSpec and identifies task-relevant information that is shared between the Specializer and the \emph{Problem Solver}. 
The output is a data structure that we call an \emph{n-tuple}. 
An n-tuple consists of nested key-value pairs that are used as a communication language between the Specializer and the Problem Solver.
N-tuples can i) specify commands, ii) represent a query, or iii) assert embedded knowledge about the world. An example excerpt of a command-type n-tuple is provided in Figure \ref{fig:ntuple}.
N-tuples are generated using templates that are shared among the ECG Analyzer, Specializer and Problem Solver, and are aligned to a shared ontology. 
The sharing of structure is crucial for the interoperation between these three parts, e.g. when it comes to generating a clarification dialog.
\begin{enumerate}
\item[\textbf{Commands}]  are to be directly executed by a robot, such as the command: \emph{``PR2, bring the soda can to the dining table!''}. 
% For example, ``TODO: language encoding of a command that is at least similar to the example scenarios'' is encoded as the following N-Tuple: 
%\todo{TODO: Display excerpt of N-Tuple}
% For example: \emph{``Robot1, dash to the green box and then amble to the blue one!''} \quad 
% Commands include the following data structure:
%\begin{tabular}{r|l|}
%\hline
%\co{predicate\_type} : &  \co{``command''} \\ \hline
% \co{return\_type} : & \co{``error\_descriptor''} \\ \hline
%\end{tabular}
\item[\textbf{Queries}]  are user questions to retrieve information about the surrounding or the robot state. For example: \emph{``Which marker is blue?''}.
% \quad 
%N-tuples for queries include the following data structure, with the \emph{return\_type} reflecting the kind of query.
%\todo{TODO: Display excerpt of N-Tuple}
%\begin{tabular}{r|l|}
%\hline
%\co{predicate\_type} : &  \co{``query''} \\ \hline
% \co{return\_type} : & \co{``boolean''} \\ \hline
%\end{tabular}
\item[\textbf{Assertions}] assign a value to a world property that is unknown to the Situation Model. For example: \emph{``The marker is under the table.''} 
Since we focus only on language understanding in this work, we assume that all assertions made by a user are correct, and neglect belief revision and related epistemic issues. 
%N-tuples for assertions include the following structure:
%\todo{TODO: Display excerpt of N-Tuple}
%\begin{tabular}{r|l|}
%\hline
%\co{predicate\_type} : &  \co{``assertion''} \\ \hline
%\co{return\_type} : & \co{``error\_descriptor''} \\ \hline
%\end{tabular}
\end{enumerate}
% The data structure depicted in Figure \ref{fig:ntuple} is a JSON visualization of the n-tuple for the command ``PR2, bring the soda can to the dining table!'' (the relevant SemSpec is shown in Figure \ref{fig:semspec}). 

Next, the \emph{Problem Solver} receives the n-tuple and determines what course of action to take.
If the user poses a question, the Problem Solver answers it; if the user orders the robot to carry out a task, the Problem Solver uses reasoning to successfully fulfill the user's request. 
This requires access to information about the world, which the Problem Solver represents as a Situation Model (see Figure \ref{fig:diagram}). 
The Situation Model ranges in complexity from the locations of objects, to spatial relations and properties of these objects.
% 
% 
% If multiple robots are involved in an application, they also communicate using n-tuples \cite{Trott2015}, and the solver then makes API calls to the underlying application, which is the CQI in this case.

\subsection{Integration of the ECG analyzer in ROS}
\label{sec:sys_desc:ROS}
To integrate the ECG analyzer software with ROS, we build two ROS packages. One is the ECG wrapper package which embeds the ECG analyzer to translate natural language into high-level robot commands. The  other one is the Command and Query Interface (CQI) which provides a robot-specific abstraction layer to translate high-level commands to low-level motor control. 

\textbf{ECG wrapper.} \quad
The ECG analyzer and solver are accessed by a ROS wrapper package which is responsible for publishing high-level commands to the CQI via topic \co{/cqi/command/}, and which receives feedback and high-level sensor data from the CQI via topic \co{/cqi/data/}. 
For now, we focus mainly on object manipulation. 
Hence, the set of commands that we currently support are moving (C1), grasping (C2) and releasing objects (C3). This set could easily be extended.
\begin{enumerate}	
	\item[\textbf{(C1)}] \co{move\_to\_pose(x, y, $\theta$)}
	\item[\textbf{(C2)}] \co{grasp\_object(object\_label)}
	\item[\textbf{(C3)}] \co{release()}
\end{enumerate}
It is the responsibility of the ECG problem solver to ensure that moving from the robot's current pose to the destination pose is possible, i.e., that there exists a feasible trajectory between start and destination pose. 
The solver also has to assure that the object label of an object is known, and that, before grasping, the robot is in a pose which makes the grasping possible (i.e., not too far away from the object and oriented towards it, and not already holding another object). 
The latter can be relaxed when the grasp action is implemented robustly, such that the robot internally fine-tunes its pose before performing the actual grasp. 

After a command is executed, ECG's problem solver expects feedback to determine whether an action was successfully finished. 
This feedback is received via messages of the form (D1) -- (D3) on ROS topic \co{cqi/data/} from the CQI, as described in the following.

\textbf{Command and Query Interface (CQI).} \quad
% We provide the ECGworkbench software as ROS module which communicates with another module that we call the \emph{Command and Query Interface} (CQI). 
The CQI is a modular interface between the low-level motor commands required by the robot hardware and the high-level action commands given by ECG's Problem Solver. 
For example, the CQI is supposed to translate a high-level \co{move\_to\_pose} command to an appropriate low-level behavior that involves locomotion, navigation and obstacle avoidance. 
So far, we have built two simple CQI modules, one for the DARwin-OP, and one for the PR2 robot, with ad-hoc solutions for navigation, object recognition and motion planning.
The fixed high-level input interface (C1) -- (C3) allows one to use the ECG framework with any robot for which a CQI module exists. 
Implementing a new CQI module is relatively simple, given the various open-source packages for moving, grasping, localization, etc. that are available for ROS.
However, our focus here is more on the interface than on the implementation of the low-level behavior, and we leave a more sophisticated implementation to the respective robotics experts. 
The interface is provided as a base class in Python, from which robot-specific CQI's inherit. 
The communication flow is implemented as usual with the ROS-internal topic-based communication paradigm, and depicted in Fig. \ref{fig:ecg-ros-commstructure}.%

The possible data messages that the CQI can currently publish are as follows:
\begin{enumerate}
	\item[\textbf{(D1)}] \co{at\_pose(x,y,$\theta$)} \quad This is a dedicated data message to communicate the robot's current position and rotation in 2d. 
	It is published continuously, and therefore inherently encodes success and failure of a \co{move\_to\_pose} action. 
	\item[\textbf{(D2)}] \co{holding(object)} \quad The message communicates which object the robot is currently holding, returning \co{none} if no object is currently held, e.g., when grasping was unsuccessful. 
	\item[\textbf{(D3)}] \co{has\_property(object, property, value)} \quad This is a more general data message that can be used to talk about object properties in general. 
	Examples are color, shape and location of objects in the environment, that the robot determines with its vision system. 
\end{enumerate}
The data interface can be extended as required by a specific scenario, to meet the capabilities of specific robots. 
Note that the robot pose and holding information could in principle also be encoded by (D3), but we prefer separate kinds of messages that use different communication properties, such as the publishing rate, which is higher for the pose and lower for the holding.

%
%!TEX root = NLU-Robots.tex

\begin{table*}[th!]\centering
\ra{1.3}
\vspace{6pt}
%\scriptsize
 \begin{tabular}{ p{0.18\textwidth} p{0.285\textwidth} 
 	p{0.018\textwidth} p{0.018\textwidth} p{0.018\textwidth} p{0.018\textwidth} 
 	p{0.018\textwidth} p{0.018\textwidth} p{0.018\textwidth}  p{0.018\textwidth} p{0.018\textwidth} p{0.018\textwidth} p{0.018\textwidth} } 
 \toprule
 Feature & Example & \multicolumn{11}{l}{Supported by} \\
  &  & [X] & \cite{Matuszek2012} & \cite{Cantrell2010} & \cite{Bastianelli2014a} & \cite{Barrett2015a} & \cite{MacMahon2006} & \cite{Steels2012} & \cite{Bos2007} & \cite{Kruijff2010} & \cite{Chai2014} & \cite{Tellex2012}\\ \midrule
 Conceptual and constructional compositionality &  
 	Abstract transitive and ditransitive constructions affect the meaning of verbs. & 
 	\checkmark & (\checkmark) & (\checkmark) & - & - & - & (\checkmark)  & (\checkmark) & (\checkmark) & (\checkmark) & - \\
 Conditionals &  
 	``If there is a can of coke on the table, please bring it!'' & 
 	\checkmark & - & - & - & - & - & (\checkmark) & (\checkmark) & - & - &  - \\
 Clarification and Dialog & 
 	``Please pick up the marker under the table!'' -- ``Which one, the red one or the blue one?'' & 
 	\checkmark & - & \checkmark & - & - & - & \checkmark & - & \checkmark &  \checkmark & \checkmark \\
Indirect assertions via relative sentences or appositions & 
	``Please bring the plate, which fell from the table, to the dishwasher.'' & 
	\checkmark & - & - & - & \checkmark & - & - & (\checkmark) & \checkmark & - & - \\
 Modalities & 
	``Are you able to order pizza?'' & 
	\checkmark & - & - & - & - & - & - & (\checkmark) & - & - & - \\
 Metaphor &   	
	``Robot, fly over here!'' (move fast)& 
   	- & - & - & - & - & - & - & - & - & - & - \\
 Indirect speech acts & 
	``Do you know where I have left my keys?'' & 
	- & - & \checkmark & - & - & - & - & - & - & - & - \\
 Sloppy or erroneous input & 
 	Erroneous input: ``Please bring me the grass'' correct input: ``Please bring me the glass'' &
 	- & - & \checkmark & - & - & - & \checkmark & - & - & - & - \\
 Disfluency analysis and repair& 
   	``There is a blue kit, erm, I mean a blue box, at the end of the table'' & 
   	- & - & \checkmark & (\checkmark) & - & - & \checkmark & - & \checkmark & - & - \\
 Verbal interlocutor feedback & 
 	``So turn right'' -- ``okay'' -- ``and walk a little bit and turn right again'' & 
   	- & - & \checkmark & - & - & - & - & - & (\checkmark) & - & - \\
 Robot implementation & 
 	 & 
   	\checkmark & - & \checkmark & - & \checkmark & (\checkmark) & \checkmark & \checkmark & - & \checkmark  & \checkmark \\
 \bottomrule
 \end{tabular}
\caption{Features of different language understanding frameworks for robots ([X] represents this work).}
\label{tab:comparisonOfFeatures}
\vspace{-10pt}
\end{table*}

\section{PROOF OF CONCEPT AND EVALUATION}
\label{sec:eval}
To provide a proof of concept, we present scenarios two indoor assistance scenarios, with additional technical detail in the supplementary video. 
We also present a qualitative comparison of the state of the art in NLU for HRI.

\subsection{Proof of Concept Scenarios}
\label{sec:eval:scenarios}
\begin{figure}
\includegraphics[width=.49\columnwidth]{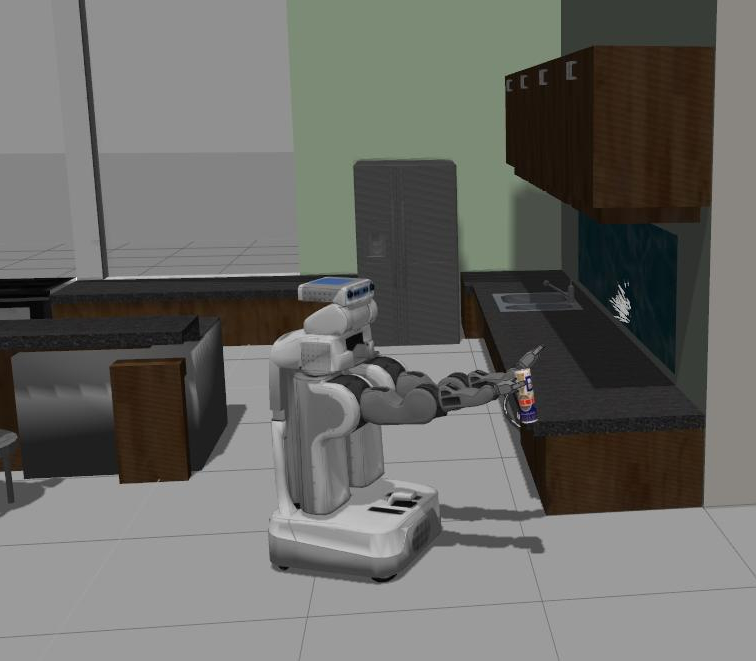}
\includegraphics[width=.49\columnwidth]{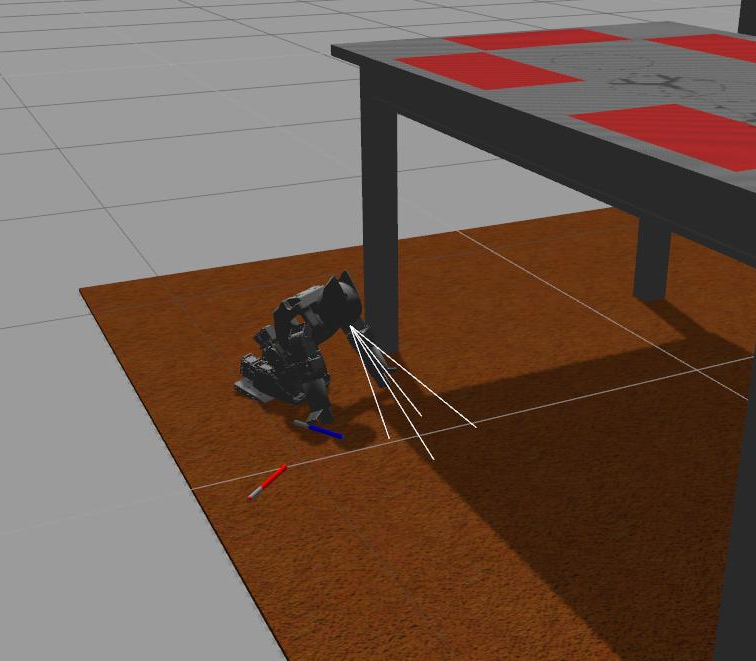}
\caption{PR2 carrying a can of soda in the kitchen environment (left) and DARwin-OP picking up the blue marker (right).}
\label{fig:demo_screenshots}
\vspace{-10pt}
\end{figure}

To demonstrate the NLU-system, we present two scenarios from the Assisted Living domain in an environment as depicted in Figure \ref{fig:demo_screenshots}. Herein, we focus on natural language problems, and assume that reasonably good speech recognition, indoor navigation and object recognition methods by respective experts are available. We currently have integrated some basic ad-hoc solutions for these problems within ECG's problem solver and the CQI. 

\textbf{Scenario 1:} \emph{Anaphora resolution and conditionals with PR2.} \quad 
The first scenario is triggered by the sentence ``PR2, if a soda can is on the kitchen counter, please bring it to the dining table, otherwise get a new one from the fridge''. Here we have a conditional if-then-else command with reference resolution. Figure \ref{fig:demo_screenshots} shows that our simulated PR2 has found the can of soda and is driving to the dining table.

\textbf{Scenario 2:} \emph{Spatial clarification dialog for reference grounding with DARwin-OP.} \quad
Here we show how our system deals with the sentence ``Darwin, pick up the marker under the table'' (see Fig. \ref{fig:demo_screenshots}). Darwin knows that there are two markers which fell down from the table, a blue and a red one. Hence, it tries to resolve this ambiguity by asking for clarification: ``Which one?''. 
In this case we answer ``The blue one'' and Darwin walks to the table to pick the blue marker up. 
Note that such clarification dialogs could also involve multiple steps. 
For example, if there were two blue markers of different sizes, Darwin would continue to resolve the ambiguity by asking for the size. 

\subsection{Comparison of Features of NLU-Systems for HRI}
\label{sec:eval:related_work}
To evaluate our work wrt. the state of the art, we look at the features of related systems found in literature and compare it to our system. 
Results are depicted in Table \ref{tab:comparisonOfFeatures}.
We not only highlight the capabilities of our system, but also looked at the literature and compiled the features that were highlighted together. Cases marked as (\checkmark) are borderline, e.g. when constructional but no conceptual compositionality is supported. These cases are explained in Section \ref{sec:prelim:survey}. 

While Table \ref{tab:comparisonOfFeatures} provides a good overview on the state of the art, it is not intended to reflect details about the depth of the individual author's focus. 
For example, \cite{Tellex2012} focus heavily on clarification and dialog, and this aspect of their work is much more sophisticated than our clarification methods.
However, we think that a successful HRI system should be able to capture as many of the listed features as possible to be successful. We also do not claim that the table is exhaustive, but we believe that it includes the vast majority of problems that occur in NLU for robotics. 

%
%!TEX root = NLU-Robots.tex
\section{CONCLUSION}
\label{sec:conclusion}
We demonstrate the importance of capturing deep semantics in language for robotic applications using examples from the indoor assistance domain. Herein, we connect the ECG analyzer tool to ROS and extend the ECG core grammar towards more robot-specific tasks. 
As a proof-of-concept, we implement two scenarios which illustrate a subset of the capabilities of the ECG analyzer. 
We note that ECG's capabilities go beyond the presented scenarios, and, for example, also allow for multi-step clarification and multi-agent communication, as shown in \cite{Khayrallah2015}. 
To realize our toolchain, we provide a modular abstraction layer (the CQI), which maps high-level actions like grasp and move to robot-specific low-level motor commands. 
So far, we have implement basic CQI modules for the PR2 and the DARwin-OP robot. 

We also present a survey on NLU for robotics, which illustrates shortcomings of our and other approaches. 
The survey supports our hypothesis: 
By building on cognitive linguistic theories, in the form of core ECG schemas and constructions, we are able to analyze a combination of several kinds of sentences that no other NLU system for robotics can currently interpret correctly and reliably. 
We demonstrate this in our scenarios, with conditional sentences in combination with anaphora resolution and grounding. 

Our approach is straight-forward to extend to other not yet supported problems that we depict in Table \ref{tab:comparisonOfFeatures}.
For example, indirect speech acts could be resolved by adding an intention detection mechanism to the problem solver, as realized by \cite{Williams2015}, and similarly for erroneous input. 
There is also ongoing work to make the ECG core grammar capable of understanding metaphor \cite{Dodge2015}. 

In addition to extending the ECG core functionality, we want to investigate how learning-based approaches for language understanding could leverage our knowledge-based approach, and vice versa. 
Specifically, we think that compositionality makes learning easier because it allows one to maintain a universal core grammar set for learning new constructions and words. Hence, by building high-level, domain-independent semantics, we can expand to new domains in a semi-automated way. 
The data that is needed for the learning can be extracted from Wizard-of-Oz experiments, but also from existing resources like FrameNet \cite{Baker1998} to build domain-specific low-level constructions and tokens that naturally compose with the core grammar.

\bibliographystyle{plain}
\bibliography{library}

\begin{thebibliography}{10}

\bibitem{Aiello2013}
Luigia~Carlucci Aiello, Emanuele Bastianelli, Luca Iocchi, Daniele Nardi,
  Vittorio Perera, and Gabriele Randelli.
\newblock {Knowledgeable talking robots}.
\newblock {\em Lecture Notes in Computer Science - AGI}, 7999:182--191, 2013.

\bibitem{Baker1998}
Collin Baker, Charles Filmore, and John Lowe.
\newblock {The Berkeley FrameNet project}.
\newblock In {\em International Conference on Computational Linguistics
  (COLING)}, 1998.

\bibitem{Barrett2015a}
Daniel~Paul Barrett, Scott~Alan Bronikowski, Haonan Yu, and Jeffrey~Mark
  Siskind.
\newblock {Robot Language Learning, Generation, and Comprehension}.
\newblock 2015.

\bibitem{Bastianelli2013}
Emanuele Bastianelli, Giuseppe Castellucci, Danielo Croce, and Roberto Basili.
\newblock {Textual Inference and Meaning Representation in Human Robot
  Interaction}.
\newblock In {\em Joint Symposium on Semantic Processing}, 2013.

\bibitem{Bastianelli2014a}
Emanuele Bastianelli, Giuseppe Castellucci, Danielo Croce, Roberto Basili,
  Daniele Nardi, and Luca Iocchi.
\newblock {HuRIC : a Human Robot Interaction Corpus}.
\newblock In {\em LREC}, 2014.

\bibitem{Bastianelli2014}
Emanuele Bastianelli, Giuseppe Castellucci, Danilo Croce, Roberto Basili, and
  Daniele Nardi.
\newblock {Effective and Robust Natural Language Understanding for Human Robot
  Interaction}.
\newblock In {\em European Conference on Artificial Intelligence}, 2014.

\bibitem{Beuls2012}
Katrien Beuls, Remi {Van Trijp}, and Pieter Wellens.
\newblock {Diagnostics and repairs in fluid construction grammar}.
\newblock In {\em Language Grounding in Robots}. Springer, 2012.

\bibitem{Bos2007}
Johan Bos and Tetsushi Oka.
\newblock {A spoken language interface with a mobile robot}.
\newblock {\em Artificial Life and Robotics}, 11(1):42--47, 2007.

\bibitem{jibo2015}
Cynthia Breazeal.
\newblock Jibo, the world's first family robot.

\bibitem{Bryant2008}
John~Edward Bryant.
\newblock {\em {Best-Fit Constructional Analysis}}.
\newblock PhD thesis, University of California at Berkeley, 2008.

\bibitem{Cantrell2010}
Rehj Cantrell, Matthias Scheutz, Paul Schermerhorn, and Xuan Wu.
\newblock {Robust spoken instruction understanding for HRI}.
\newblock In {\em International Conference on Human-Robot Interaction (HRI)},
  2010.

\bibitem{Chai2014}
Joyce~Y Chai, Lanbo She, Rui Fang, Spencer Ottarson, Cody Littley, Changsong
  Liu, and Kenneth Hanson.
\newblock {Collaborative effort towards common ground in situated human-robot
  dialogue}.
\newblock In {\em International Confernce on Human-Robot-Interaction (HRI)},
  2014.

\bibitem{Deits2012}
Robin Deits, Stefanie Tellex, Pratiksha Thaker, Dimitar Simeonov, Thomas
  Kollar, and Nicholas Roy.
\newblock {Clarifying Commands with information-Theoretic Human-Robot Dialog}.
\newblock {\em Journal of Human-Robot Interaction}, 2012.

\bibitem{Dodge2015}
Ellen Dodge, Jisup Hong, and Elise Stickles.
\newblock {MetaNet : Deep semantic automatic metaphor analysis}.
\newblock In {\em Workshop on Metaphor in NLP, at NAACL}, 2015.

\bibitem{Eppe2013c}
Manfred Eppe, Mehul Bhatt, and Frank Dylla.
\newblock {Approximate Epistemic Planning with Postdiction as Answer-Set
  Programming}.
\newblock In {\em International Conference on Logic Programming and
  Nonmonotonic Reasoning (LPNMR)}, 2013.

\bibitem{Feldman2009}
Jerome Feldman, John~Edward Bryant, and E~Dodge.
\newblock {Embodied Construction Grammar}.
\newblock In {\em The Oxfrod Handbook of Computational Linguistics}, pages 38
  -- 111. Oxford University Press, 2009.

\bibitem{Feldman2009a}
Jerome Feldman, Ellen Dodge, and John Bryant.
\newblock {A neural theory of language and embodied construction grammar}.
\newblock {\em The Oxford Handbook of Linguistic Analysis.}, pages 111 ----
  138, 2009.

\bibitem{Fillmore1985}
Charles Fillmore.
\newblock {Frames and the semantics of understanding}.
\newblock {\em Quaderni di Semantica}, 6(2):222--254, 1985.

\bibitem{ros2015}
Open Source~Robotics Foundation.
\newblock Ros.org.

\bibitem{Goldberg1995}
Adele Goldberg.
\newblock {\em {Constructions: A Construction Grammar Approach to Argument
  Structure}}.
\newblock University of Chicago Press, 1995.

\bibitem{Kamp2011}
Hans Kamp, Josef van Genabith, and Uwe Reyle.
\newblock {Discourse Representation Theory}.
\newblock In {\em Handbook of Philosophical Logic}, pages 125 -- 394. 2011.

\bibitem{Khayrallah2015}
Huda Khayrallah, Sean Trott, and Jerome Feldman.
\newblock {Natural Language For Human Robot Interaction}.
\newblock In {\em International Conference on Human-Robot Interaction (HRI)},
  2015.

\bibitem{Kruijff2010}
Geert-Jan Kruijff, Pierre Lison, Trevor Benjamin, Henrik Jacobsen, Hendrik
  Zender, and Ivana Kruijff-Korbayov{\'{a}}.
\newblock {Situated dialogue processing for human-robot interaction}.
\newblock {\em Cognitive Systems}, 2010.

\bibitem{Kruijff2007}
Geert~Jan Kruijff, Hendrik Zender, Patric Jensfelt, and Henrik Christensen.
\newblock {Situated dialogue and spatial organization: What, where... and why?}
\newblock {\em International Journal of Advanced Robotic Systems}, 2007.

\bibitem{Lakoff1980}
George Lakoff and Mark Johnson.
\newblock {\em Metaphors We Live By}.
\newblock University of Chicago Press, 1980.

\bibitem{Lakoff1999}
George Lakoff and Mark Johnson.
\newblock {\em {Philosophy in the Flesh}}.
\newblock Basic Books, 1999.

\bibitem{MacMahon2006}
M~MacMahon, Brian Stankiewicz, and Benjamin Kuipers.
\newblock {Walk the talk: Connecting language, knowledge, and action in route
  instructions}.
\newblock In {\em AAAI}, 2006.

\bibitem{Matuszek2012}
Cynthia Matuszek, Evan Herbst, Luke Zettlemoyer, and Dieter Fox.
\newblock {Learning to parse natural language commands to a robot control
  system}.
\newblock {\em Int'l Symposium on Experimental Robotics (ISER)}, 2012.

\bibitem{Petruck1996}
Miriam Petruck.
\newblock {Frame Semantics}.
\newblock In {\em Handbook of Pragmatics}. John Benjamin Publishing Company,
  1996.

\bibitem{Pullum1982}
Geoffrey Pullum and Gerald Gazadar.
\newblock {Natural languages and context-free languages}.
\newblock {\em Linguistics and Philosophy}, 1982.

\bibitem{Scheutz2013}
Matthias Scheutz, Gordon Briggs, Rehj Cantrell, Evan Krause, Tom Williams, and
  Richard Veale.
\newblock {Novel mechanisms for natural human-robot interactions in the diarc
  architecture}.
\newblock In {\em AAAI}, 2013.

\bibitem{Scheutz2008}
Matthias Scheutz and Kathleen Eberhard.
\newblock {Towards a framework for integrated natural language processing
  architectures for social robots}.
\newblock In {\em International Workshop on Natural Language Processing and
  Cognitive Science}, 2008.

\bibitem{Sinha2008}
Steve Sinha.
\newblock {\em {Answering Questions about Complex Events}}.
\newblock PhD thesis, University of California at Berkeley, 2008.

\bibitem{Spranger2015}
Michael Spranger and Luc Steels.
\newblock {Co-Acquisition of Syntax and Semantics — An Investigation in
  Spatial Language}.
\newblock In {\em International Joint Conference on Artificial Intelligence},
  2015.

\bibitem{Steedman2000}
Mark Steedman.
\newblock {\em {The syntactic process}}.
\newblock MIT Press, 2000.

\bibitem{Steels1998a}
Luc Steels.
\newblock {The origins of ontologies and communication conventions in
  multi-agent systems}.
\newblock {\em Autonomous Agents and Multi-Agent Systems}, 1(2):169--194, 1998.

\bibitem{Steels2015}
Luc Steels.
\newblock {\em {The Talking Heads Experiment}}.
\newblock 2015.

\bibitem{Steels2012}
Luc Steels, Joachim {De Beule}, and Pieter Wellens.
\newblock {Fluid Construction Grammar on Real Robots}.
\newblock In {\em Language Grounding in Robotics}. Springer, 2012.

\bibitem{Tellex2011}
Stefanie Tellex, Thomas Kollar, Steven Dickerson, Matthew~R Walter, Ashis~Gopal
  Banerjee, Seth Teller, and Nicholas Roy.
\newblock {Understanding natural language commands for robotic navigation and
  mobile manipulation}.
\newblock In {\em AAAI Conference on Artificial Intelligence}, 2011.

\bibitem{Tellex2012}
Stefanie Tellex, Pratiksha Thaker, Robin L~H Deits, Dimitar Simeonov, Thomas
  Kollar, and Nicholas Roy.
\newblock {Toward Information Theoretic Human-Robot Dialog}.
\newblock {\em Robotics: Science and Systems Conference}, 2012.

\bibitem{Trott2015}
Sean Trott, Aur{\'{e}}lien Appriou, Jerome Feldman, and Adam Janin.
\newblock {Natural Language Understanding and Communication for Multi-Agent
  Systems}.
\newblock In {\em AAAI Fall Symposium}, pages 137--141, 2015.

\bibitem{Walter2013}
Matthew~R Walter, Sachithra Hemachandra, Bianca Homberg, Stefanie Tellex, and
  Seth Teller.
\newblock {Learning Semantic Maps from Natural Language Descriptions}.
\newblock {\em Robotics Science and Systems}, pages 1--8, 2013.

\bibitem{Williams2015}
Tom Williams, Gordon Briggs, Bradley Oosterveld, and Matthias Scheutz.
\newblock {Going Beyond Literal Command-Based Instructions : Extending Robotic
  Natural Language Interaction Capabilities}.
\newblock In {\em AAAI}, 2015.

\bibitem{Wisspeintner2009}
Thomas Wisspeintner, Tijn van~der Zant, Luca Iocchi, and Stefan Schiffer.
\newblock {RoboCup@Home: Scientific Competition and Benchmarking for Domestic
  Service Robots}.
\newblock {\em Interaction Studies}, 10(3):392--426, 2009.

\end{thebibliography}
\end{document}